# Optimizing Offensive Gameplan in the National Basketball Association with Machine Learning

Eamon Mukhopadhyay, Purdue University

*Abstract*—Throughout the analytical revolution that has occurred in the NBA, the development of specific metrics and formulas has given teams, coaches, and players a new way to see the game. However - the question arises - how can we verify any metrics? One method would simply be eyeball approximation (trying out many different gameplans) and/or trial and error - an estimation-based and costly approach. Another approach is to try to model already existing metrics with a unique set of features using machine learning techniques. The key to this approach is that with these features that are selected, we can try to gauge the effectiveness of these features combined, rather than using individual analysis in simple metric evaluation. If we have an accurate model, it can particularly help us determine the specifics of gameplan execution. In this paper, the statistic ORTG (Offensive Rating, developed by Dean Oliver) was found to have a correlation with different NBA playtypes using both a linear regression model and a neural network regression model, although ultimately, a neural network worked slightly better than linear regression. Using the accuracy of the models as a justification, the next step was to optimize the output of the model with test examples, which would demonstrate the combination of features to best achieve a highly functioning offense.

I. INTRODUCTION

1.1 Context of the Paper

The NBA (National Basketball Association) distinguishes itself from many other sports leagues in the world due to the vastness of its specific metrics that go beyond the general box score quantities. Some of these metrics have been manufactured based upon raw statistics, while other non-box-score metrics represent raw statistics themselves. Two notable examples of manufactured statistics include Player Efficiency Rating (PER) and Usage Rate (USGR), developed by John Hollinger. In this context, raw statistics can be defined as statistics that are not built upon other statistics. For example, field goal attempts (FGA) are collected upon iterative tracking - there is no formula to calculate FGA in the NBA.

There were some starting points - notably shown by the revolution in basketball analytics since the early 2000's; Basketball on Paper (Oliver, 2004) by Dean Oliver [8], was a notable pioneering work to guide people towards using metrics, in addition to the eye-test, to make basketball decisions. Oliver's works were heavily centered around building metrics which could show reasonable contrasts based on team performance; these contrasts were verified mathematically and using the eye-test. The developed metrics are extremely valuable for their ability to combine multiple other metrics using a reasonable method.

Determined from analyzing ORTG (Offensive Rating), developed by Oliver, it is known that increasing a team's points scored will increase their ORTG.

$$100 \cdot \frac{(Points\ Scored)}{(Possessions)} = TeamORTG$$

However, it is well understood the main goal of winning basketball (offensively) is to score points in some way on a given possession. This raises the question: how do we determine what truly derives ORTG? Simply put, how did a team increase their offensive output?

As many would likely predict, the answer is slightly complicated. But one key revolution to help us answer this question has been the expansion of specific datasets which are able to be accessed by any person who can access the internet. Organizations such as Second Spectrum and Synergy have provided efforts to display unique raw analytics, such as individual player isolation frequency.

Also, the expansion of machine learning techniques has provided basketball analysts a method to create functions which can model trends in basketball. As such, this research paper strongly consisted of trying different models, along with data condensing methods to truly determine how to mathematically optimize team gameplan.

1.2 Literature Review

There has been plenty of research regarding offensive schemes, specifically shooting. Huangchang Gou and Hui Zhang [4] found a negative correlation between two-point field goal ratio and winning probability, but a positive correlation between three-point field goal ratio and winning probability (Gou and Zhang, 2022). Konstantinos Pelechrinis and Kirk Goldsberry (notably former vice president for strategic research for the San Antonio Spurs) [9] also used k-means clustering to determine that half of corner three pointers are caused by shooter waiting in the corner for a pass (Pelechrinis and Goldsberry, 2021).

In discussions regarding how to apply mathematics and computing to basketball, there are multiple different formulas being proposed for new metrics. As Kubatko et al. [5] show, some of these formulas are built for the purposes of determining a percentage or per minute statistic, such as the "Per-40-Minute" statistic, simply calculated by multiplying a per-game statistic by a ratio defined as [40 divided by player minutes per game]. Other formulas are built as estimations from personal basketball knowledge. For instance, EFF (efficiency) is an estimation built upon some common statistics including points, rebounds, assists, etc. (Kubatko, 2007).

Machine learning has been incorporated to an extent in basketball analysis and research, but not to a very large extent. Much individual basketball machine-learning research has been performed with the purpose of determining real-time predictions. Classification (discrete modeling) has been an important part of modeling, through examples like win/loss and make/miss. Cameron Fuqua [3], over a six year period of twenty four NCAA final four teams, was able to predict fourteen of the final four teams using a logistic regression model (Fuqua, 2014). Also, Shah and Romijnders [11] found that using a recurrent neural network with positional values and game clock variables could accurately model three pointer success at a specific distance to the basket (Shah and Romijnders, 2016).

Regression methods have also been commonly used to model trends in the NBA. Ciampolini et al. [2] argue for a lack of association between passing and shooting efficacy, but a large association between shooting condition and shooting efficiency using simple regression algorithms (Ciampolini, 2018). Supola et al. [10] and Ulas [12] also demonstrate this, with Supola et. al using quantile/linear regression to model variables associated with the "extra pass" (Supola et al., 2023) and Ulas using linear regression to model team value based on a variety of features (Ulas, 2021). When used in a traditional setting, Nguyen [6] found lower model accuracy to predict NBA player performance and accuracy using a DNN (deep neural network) rather than traditional machine learning methods (Nguyen, 2021).

On a somewhat interesting note, some of the past approaches of attempting to apply neural networks to basketball have involved some type of common application of a neural network. For instance, Oved [7] has attempted to model game-by-game performances from interviews (natural language processing) and other features using an LSTM and linear classification (Oved et al., 2020).

Cabarkapa et al. [1] have already concluded statistically significant differences in multiple recorded box-score statistics (e.g. field goals made and attempted) between the regular season and playoffs (Cabarkapa et al., 2022).

### 1.3 Purpose of this Paper

This paper intends to develop multiple hypotheses based upon a model which incorporates features based upon the playtype of a possession to predict the dependent variable "Offensive Rating". These models will help get a glimpse of the effectiveness of each feature to maximize team gameplan.

The first main goal of this paper is to determine the machine learning models which best fit the data, and if a high enough model-accuracy occurs, display hypothetical testing data which may optimize a team's ORTG (which, in turn, may represent a specific offensive gameplan to be used as well). Playtype prioritization can be reflected in team gameplan through coach instruction and player tendencies.

The second main goal of this paper is to evaluate the effectiveness of a heavy emphasis on advanced features to optimize a model (which can accurately predict a team's ORTG), rather than simply creating a new metric. One unique part about this study is that none of the features in any model will use any box-score statistics. Team gameplan optimization goes far beyond recorded points, rebounds, turnovers, etc. but few basketball statistical models truly ignore them. However, the dependent variable, ORTG, is built upon multiple box score statistics - deriving (rather than confirming) how ORTG is built is very important to identifying team gameplan.

## II. METHODOLOGY

### 2.1 Study Design and Tools

All of the data acquired from this work is from the statistics section of the official NBA website (https://www.nba.com/stats). Player playtype statistics are recorded by Synergy. This study will attempt to model offensive rating by means of two machine learning procedures: linear regression and a supervised, multi-layer perceptron neural network.

The benefit towards using ORTG as the dependent variable (rather than simply points scored) is that it accounts for how many possessions a team had to score those points. This is an important scaling device which can help analyze different years of basketball (the average number of possessions per team differs on a year-by-year basis).

There was no relevant ethical consideration, nor review needed because this paper analyzes public data provided to everyone with internet access.

### 2.2 Initial Feature Selections and Justification

As mentioned, this study will model offensive playtype against ORTG; the feature branches will be from the playtypes Isolation, Transition, Pick and Roll Ball Handler, Pick and Roll Roll Man, Post Up, Spot Up, Cut and Off Screen. The playtypes "Putbacks", "Misc" (Miscellaneous), and "Handoff" were not included because they, more or less, do not represent a specific offensive scheme or gameplan (as mentioned in 1.3 Purpose of this Paper, optimizing offensive gameplan is a major purpose of this paper). Of the listed branches, each of them consist of the same internal features - FREQ% (Frequency Percentage), FG% (Field Goal Percentage), FT% (Free Throw Frequency Percentage), TOV% (Turnover

Percentage) FREQ%, AND ONE FREQ%, and SCORE FREQ%. Some available features were excluded - for example, POSS (Possessions) and FREQ% would both point towards a frequency of a specific gameplan, but POSS may also be influenced by factors such as pace, speed, etc. Therefore, FREQ% was instead used. The total number of features for the model is 48, denoted by the eight different play types multiplied by the six internal features.

**2.3 Dataset Selections and Justifications**

The data used in the model will be within the range of the 2015-16 NBA season to the 2022-23 NBA season. The 2015-16 season represents the first year on NBA.com when playtype statistics were available.

The model will simply take the cumulative playtype statistics of every NBA team from the 2015-16 NBA season to the 2022-23 NBA season, making 240 total data points. Individual game-by-game datasets were not available, so cumulative team statistics were used.

**2.4 Machine Learning Selections and Justifications**

This study is going to test linear regression and a supervised MLP regression neural network, and determine which one works the best. In addition, rather than using a specific set number of testing samples, there are multiple occurring iterations of each model being built (this is often referred to as leave-one-out cross validation), where one different testing point is used every time, making each model slightly different. Principal component analysis was used to reduce the number of features in a model with the main goal of avoiding underfitting. Both of these tools can help to improve model accuracy, particularly when the number of samples may be limited.

The linear regression model is going to work in a straightforward manner - there are going to be a range of features (normalized) which are attached to weights with the incorporation of a bias term. ORTG will be normalized from a value 0 to 1.

$$y = w_1 x_1 + w_2 x_2 + w_3 x_3 + \ldots + b$$

The supervised MLP neural network will run on a regression-type scheme. There are going to be a range of features (normalized) then attached to weights and biases. The activation functions for each layer excluding the output layer will be ReLU (Rectified Linear Unit). There will be one target output for the model: ORTG. ORTG will also be normalized from a value 0 to 1.

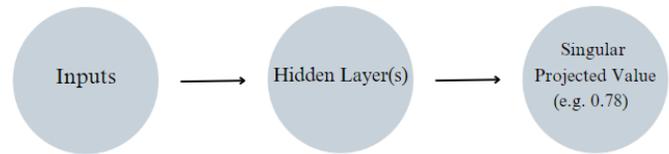

### III. ANALYSIS

**3.1 Introduction**

In the analysis section, each model was generated on Python software. The individual errors of each test sample in the different model iterations were used to calculate the RMSE (Root Mean Square Error). The computer also displayed the coefficient of determination ($R^2$) of the dataset.

**3.2 Playtype and Offensive Rating**

Upon cutting down the feature count to 18, the analysis was performed. Using simple linear regression, the calculated RMSE was found to be approximately 0.112. When adjusted to the ORTG scale, the RMSE was approximately 2.26 ORTG points. The coefficient of determination was found to be approximately 0.665, demonstrating a very strong correlation.

Using a multi-layer perceptron regressor neural network (also with a feature count of 18), it was found that the optimal hidden layer count was 1, with the size of the lone hidden layer being 3. In this model, the RMSE was calculated to be approximately 0.107; adjusted to the ORTG scale, the calculated RMSE was approximately 2.16 ORTG points. Here, the coefficient of determination was, as expected by the lower RMSE value, higher at 0.694.

Shown below is the predicted against actual values for both models.

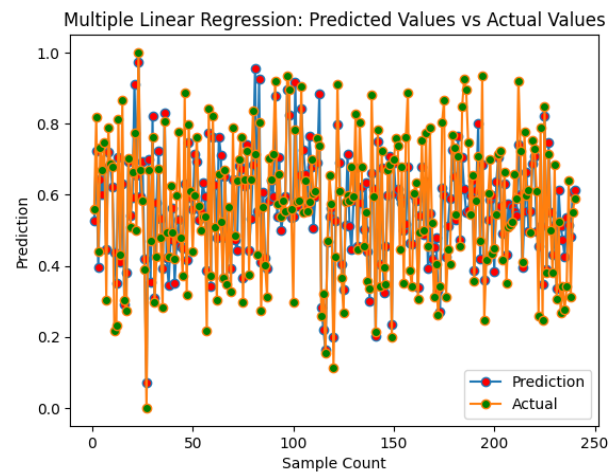

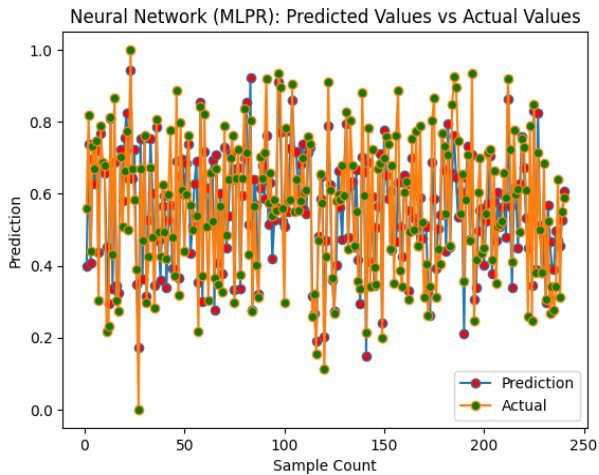

Both prediction sets look similar, which makes sense due to the fact that the optimal neural network did not seem to show a high complexity of nonlinearity. The linear model seemed to handle some of the polarized data points better than the neural network; this also makes sense considering the nature of training each model.

**3.3 Discussion of Models**

Both models provide acceptable methods to model playtype against ORTG. There are multiple instances of the applicability of both models. If a prediction of a higher-ranged ORTG is valued in a particular instance, then it can be used. The neural network performs very well in predicting intermediate values.

Though mathematically-verified for reasonably strong accuracy, these models are also supported by intuition - there are only so many possible "playtypes" in basketball. Offensive possessions are built upon these playtypes.

IV. GAMEPLAN OPTIMIZATION

**4.1 Introduction**
This section of the paper will focus on optimizing team gameplan by the method of realistic mathematical optimization of a set of inputs to a model.

**4.2 The Effective Features**

Because the neural network was the most effective model for most of the test samples, it will be used to determine the effective features when inputted. A general trend will be used to create gameplan hypotheses, rather than examples that may yield an ORTG near one or zero. Regardless, both the neural network and linear regression models will have similar conclusions.

Based upon the model, it was clear which playtypes the model encouraged to partake in most commonly: isolation, spot-up shooting, and transition.

Following up on this, the model suggested the most essential playtypes to score on and capitalize - the pick-and-roll (with the ball handler), isolation, and spot-up shooting. Also - the model encouraged pick-and-roll ball-handler, isolation, and post-up ball efficiency (i.e. not committing many turnovers) towards helping build a strong offensive rating.

**4.3 Interpretation and Gameplan Optimization**

One key point that should be noted when finding effective features and developing gameplan hypotheses is that general basketball knowledge (and an intuitive sense) should be combined with the mathematical model to develop gameplans.

The first gameplan hypothesis that can be made from the model results is that provided a team has players who have experience and reasonable ability in these types of plays, (i) one-on-one (isolation) situations should be heavily emphasized on a play-by-play basis in the NBA. Specifically, isolation plays should account for nearly 20-25% of play frequency combined. While many teams have begun using their star players in isolation situations, the goal would be to run enough one-on-one plays to be at the top of the team dataset in terms of playtype frequency. In an intuitive sense, the value behind these types of plays is that they help keep the defense off-guard throughout the game (forcing to use one-on-one ability versus predicting a play) and can help generate spot-up shooting opportunities with kick-outs due to the nature of these plays.

Expanding on the first gameplan hypothesis, it should not be understated how important efficiency is in one-on-one possessions, which is why forcing this type of gameplan on a lineup whose players have not shown a capability to win in one-on-one situations may not be very effective. It is also very important to note that this paper is delegating an in-depth analysis of whether or not the players of a team are sufficient to meet this goal to other papers. In a general sense, the top isolation players in regards to frequency, field goal percentage, and score frequency can be viewed on NBA.com (individual playtype statistics, also provided by Synergy).

The second gameplan hypothesis is that (ii) a team can maximize their offensive output by increasing their spot-up plays at both a high-efficiency and high-rate (40-42% field goal percentage or higher and 25-28% frequency is optimal) through natural, non-drawn-up offensive plays. There is a general consensus that spacing is important in the NBA, but encouraging kick-outs, rather than forced shots, from one-on-one plays or acquiring points through the context of a normal possession (these include drives, pull-ups, floaters which are counted as spot-up possessions in the dataset) without too many drawn-up plays can help build a strong offensive foundation.

The third gameplan hypothesis is that (iii) transition opportunities should be used to help generate offensive firepower, at around 17% to 20% of general plays. To note, the true impact this scheme may have on other factors like defensive intensity and offensive discipline is not being measured in this study.

Finally, the fourth hypothesis is that (iv) an efficient, non-aggressive scoring approach to the pick and roll may generally yield higher offensive output. Also, pick and roll ball efficiency is, as mentioned above, important and should be heavily emphasized in a gameplan. The best scoring methods for teams are spot-up, transition, and isolation plays. The gameplan would include pick-and-roll possessions finished (defined as ending with a field goal attempt or turnover) by the ball-handler and roll-man being utilized at a general rate of around 15% frequency. In an intuitive sense, pick-and-roll plays can help force defensive help, which demonstrates the value of this method - to kick-out to open shooters and provide additional spot-up opportunities.

V. CONSIDERATIONS

**5.1 Exceptions**
The stated hypotheses are given by general offensive trends among teams. There will always be exceptions - for example, a team may be able to record a high offensive rating while utilizing off-ball screen plays and limiting one-on-one possessions. The Golden State Warriors have followed this trend in some of the past seasons (specifically, 2017-18, 2018-19, and 2022-23 NBA seasons), ranking highly in ORTG with a fairly unique playstyle.

This brings about an important point: the benefit of the findings from the two models is that they are general, i.e. they have multiple tested examples of accuracy. In this paper, finding an accurate model was important. It should also be considered that a team may be able to rank highly in only specific feature categories and still maintain a high ORTG with respect to teams in the previous five years). In theoretical terms, multi linear regression and neural network models combine features in such a way to optimize the respective model, but it does not necessarily mean that one specific set of inputs will yield one separate solution. For example, the 2022-23 Sacramento Kings set the NBA record for highest ORTG by a team at 118.6, but rarely utilized isolation plays. Instead, they emphasized transition and spot-up plays, which were noted in the model to be important factors towards ORTG.

**5.2 Limitations**

This study successfully mapped out four offensive gameplan hypotheses using a fairly reliable modeling system. There are, however, some limitations.

First, the size of the first dataset, and its use of cumulative statistics (rather than a game-by-game log), especially compared with the number of features, may raise questions. Though, to note, there have been measures to prevent this in the study - such as using PCA and cross-validation methods. Still, the model's reliability could have been further boosted with a larger overall dataset.

Also, the usage of higher, more advanced systems were not used. They were considered, but due to the success of the linear regression and neural network models, their respective findings were put forth without additional model testing. The neural network and linear regression models already had notable correlations, but different and/or advanced methods (bayesian methods, polynomial regression, etc.) were not utilized to determine whether or not a higher correlation could be found.

V. CONCLUSION

This paper had two main goals - first: to interpret the results of an NBA machine learning model to optimize team gameplan, and second: to determine the effectiveness of using machine learning models that determine what can build an existing metric (rather than simply creating a new one) and if successful, persuade sports analysts to use this method more commonly rather than simple metric analytics.

The first goal was fairly successful, and four hypotheses were able to be created based on the overall models, which were determined to be reliable. Essentially, some specific features were shown to have a noticeable impact on ORTG, which can be maximized to predict a higher ORTG.

The second goal has been met in its first criteria - a machine learning model was quite effective in modeling a well-known statistic. Its second criteria of persuading a more widespread use of machine learning in sports analytics cannot be fulfilled at the time of this writing, but it is in this paper's interest that the second goal is fully met in the near future.

Machine learning may not always work in every scenario, and in those scenarios simply analyzing metrics on their own may help to gain insight towards a specific team gameplans. But if a model is determined to show a reliable association between the features and output, it is possible to mathematically visualize a hypothetical gameplan using the model, as that is in the nature of prediction from machine learning models. Rather than simply using singular metrics, machine learning can allow for another tool which uses mathematics and past data examples that can be compared with normal basketball intuition.

The findings of this paper displayed some significant conclusions, which can help guide teams towards a successful offensive scheme through gameplan strategies. These can be considered by coaches, team executives, league executives, players, etc. and can potentially change the general approach to the game.

The data-plotting graphs implemented in this paper were built with the help of Matplotlib, a Python framework. The requested citation is as follows:

J. D. Hunter, "Matplotlib: A 2D Graphics Environment", Computing in Science & Engineering, vol. 9, no. 3, pp. 90-95, 2007.

REFERENCES

[1] Cabarkapa, D. V., Deane, M. A., Fry, A. C., Jones, G. T., Philipp, N. M., & Yu, D. (2022, August 19). Game Statistics that discriminate winning and losing at the NBA level of basketball competition. *PLOS One*. https://journals.plos.org/plosone/article?id=10.1371/journal.pone.0273427

[2] Ciampolini, V., Ibáñez, S. J., Nunes, E. L. G., Borgatto, A. F., & Nascimento, J. V. do. (2018, January 8). Factors associated with basketball field goals made in the 2014 NBA finals. *Motriz: Revista de Educação Física*. https://www.scielo.br/j/motriz/a/y5SJ8DWRb7XrKDzHrk84r3v/abstract/?lang=en#

[3] Fuqua, C. (2014, Spring). The Final Four Formula: A Binary Choice Logit Model to Predict the Semifinalists of the NCAA Division I Men's Basketball Tournament. *UNI ScholarWorks*. https://scholarworks.uni.edu/mtie/vol16/iss1/5/

[4] Gou, H., Zhang, H. (2022, August 23). Better Offensive Strategy in Basketball: A Two-Point or a Three-Point Shot?. *Journal of Human Kinetics*. https://sciendo.com/article/10.2478/hukin-2022-0061

[5] Kubatko, J., Oliver, D., Pelton, K., & Rosenbaum, D. T. (2007, July 9). A Starting Point for Analyzing Basketball Statistics. *Journal of Quantitative Analysis in Sports*. https://www.degruyter.com/document/doi/10.2202/1559-0410.1070/html?lang=en

[6] Nguyen, N.H., Nguyen, D.T.A., Ma, B., & Hu, J.. (2022). The application of machine learning and deep learning in sport: predicting NBA players' performance and popularity. J*ournal of Information and Telecommunication, 6(2)*. https://doi.org/10.1080/24751839.2021.1977066

[7] Oved, N., Feder, A., & Reichart, R. (2020). Predicting In-Game Actions from Interviews of NBA Players. *MIT Press*. https://direct.mit.edu/coli/article/46/3/667/93377/Predicting-In-Game-Actions-from-Interviews-of-NBA

[8] Oliver, D. (2011). *Basketball on Paper: Rules and Tools For Performance Analysis*. Potomac Books, Inc.

[9] Pelechrinis, K., & Goldsberry, K. (2021, May 26). The Anatomy of Corner 3s in the NBA: What makes them efficient, how are they generated and how can defenses respond?. *arXiv.org*. https://arxiv.org/abs/2105.12785

[10] Shah, R., & Romijnders, R. (2016, August 16). Applying Deep Learning to Basketball Trajectories. *arXiv.org*. https://arxiv.org/abs/1608.03793

[11] Supola, B., Hoch, T., & Baca, A. (2023, March 1). Modeling the extra pass in basketball – an assessment of one of the most crucial skills for creating great ball movement. *International Journal of Computer Science in Sport*. https://sciendo.com/article/10.2478/ijcss-2023-0002

[12] Ulas, E. (2021, June 17). Examination of National Basketball Association (NBA) team values based on dynamic linear mixed models. *PloS One*. https://www.ncbi.nlm.nih.gov/pmc/articles/PMC8211228/

All data and statistics were acquired from NBA.com and presented by Synergy from https://www.nba.com/stats.